\documentclass{article}

\usepackage{PRIMEarxiv}

\usepackage[utf8]{inputenc} 
\usepackage[T1]{fontenc}    
\usepackage{hyperref}       
\usepackage{url}            
\usepackage{booktabs}       
\usepackage{amsfonts}       
\usepackage{nicefrac}       
\usepackage{microtype}      
\usepackage{lipsum}
\usepackage{fancyhdr}       
\usepackage{graphicx}       
\graphicspath{{media/}}     
\usepackage{algorithm}
\usepackage{algpseudocode}
\usepackage{amsmath}
\title{Technical Report: The Graph Spectral Token -- Enhancing Graph Transformers with Spectral Information
}

\author{
  Zihan Pengmei \\
  Department of Chemistry \\
  The University of Chicago \\
  Chicago,IL,USA\\
  \texttt{zpengmei@uchicago.edu} \\
   \And
  Zimu Li \\
  Yau Mathematical Science Center \\
  Tsinghua University \\
  Beijing, China\\
  \texttt{lizm@mail.sustech.edu.cn} \\
}

\begin{document}
\maketitle

\begin{abstract}
Graph Transformers have emerged as a powerful alternative to Message-Passing Graph Neural Networks (MP-GNNs) to address limitations such as over-squashing of information exchange. However, incorporating graph inductive bias into transformer architectures remains a significant challenge. In this report, we propose \emph{the Graph Spectral Token}, a novel approach to directly encode graph spectral information, which captures the global structure of the graph, into the transformer architecture. By parameterizing the auxiliary [CLS] token and leaving other tokens representing graph nodes, our method seamlessly integrates spectral information into the learning process. We benchmark the effectiveness of our approach by enhancing two existing graph transformers, GraphTrans and SubFormer. The improved GraphTrans, dubbed GraphTrans-Spec, achieves over 10\% improvements on large graph benchmark datasets while maintaining efficiency comparable to MP-GNNs. SubFormer-Spec demonstrates strong performance across various datasets. The code for our implementation is available at \url{https://github.com/zpengmei/SubFormer-Spec}.
\end{abstract}


\section{Introduction}
Graph transformers have demonstrated impressive results compared to conventional Message-Passing Graph Neural Networks (MP-GNNs) in various graph benchmarks. They aim to solve inherent limitations of MP-GNNs, such as the over compression of information, where the recursive neighborhood aggregation can lead to loss of local information, and the under-reaching problem, where the receptive field of nodes is limited by the number of layers \cite{GraphTransformer2020,wu2021representing,SAN2021,kim2022pure,pengmei2023transformers}. The self-attention mechanism in graph transformers works as a fully-connected graph neural network, allowing for more efficient information exchange. GraphTrans \cite{wu2021representing} and SubFormer \cite{pengmei2023transformers} are two similar graph transformer architectures that combine shallow MP-GNN layers for local feature extraction and standard Transformer blocks for global information exchange. However, SubFormer incorporates the molecular coarse-graining assumption \cite{JunctionTree2018,HIMP2020}, which simplifies the graph structure by grouping nodes into substructures, while GraphTrans does not. SubFormer demonstrates that by incorporating proper prior knowledge, such as the coarse-graining assumption, satisfactory performance can be achieved without further complicating the updating function.

In this report, we propose \emph{the Graph Spectral Token} as a general method to include graph spectrum information, which captures the global structure of the graph, into the design of graph transformers. Starting from BERT \cite{devlin2018bert}, an auxiliary [CLS] token has been introduced as a global learnable pooling function from all the other tokens \cite{dosovitskiy2020image}. Conventionally, the [CLS] token is initialized with random parameters. Instead, we propose to encode the graph spectral information via the [CLS] token. Updating both spectral and ordinary graph features simultaneously through Transformer further enhances the model's expressive power than simply utilizing graph spectrum via graph convolution or its generalization \cite{henaff2015deep,bo2023specformer}. 
We extensively benchmark the improved graph transformers, termed as SubFormer-Spec and GraphTrans-Spec on multiple molecular modeling datasets. The impressive results suggest the potential to further investigate \emph{the Graph Spectral Token} as an effective and efficient way to inject graph spectral information into graph transformers.

\section{Incorporation into Graph Transformers}
\label{sec:spec_token}

\begin{figure}
    \centering
    \includegraphics[width=0.9\textwidth]{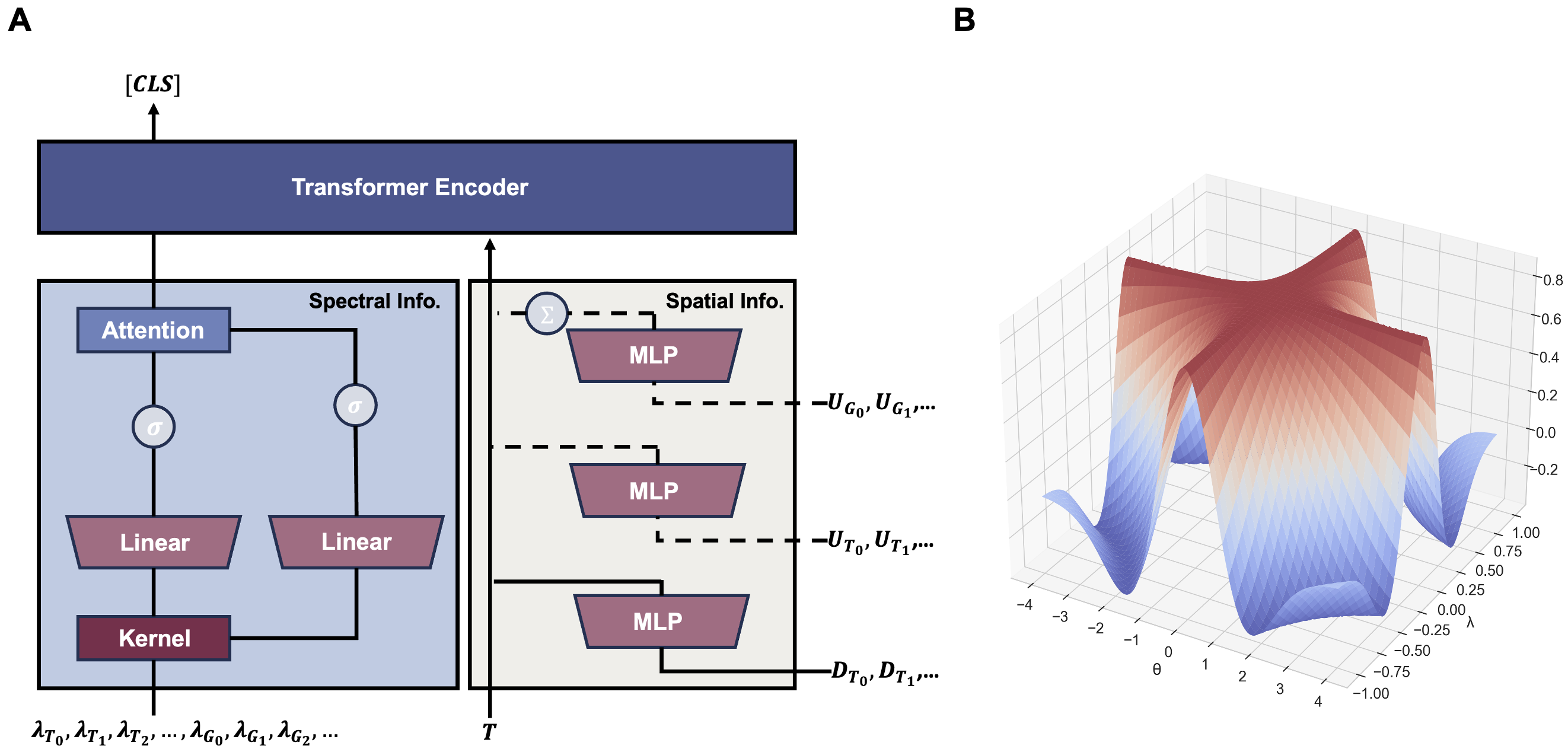}
    \caption{Illustration of the proposed \emph{the Graph Spectral Token} for graph Transformer. (A) Spectral information is processed with an auxiliary network and assigned to the [CLS] token, while ordinary node features are processed through conventional tokens with node degree and optional Laplacian eigenvectors information. (B) Visualization of Mexican Hat kernel along time constants $\theta$ and eigenvalues $\lambda$.}
    \label{fig:spec_token}
\end{figure}

\subsection{The Graph Spectral Token}

To begin with, we map eigenvalues into a higher-dimensional space which is expected to create new representations that capture more complex patterns or relationships that may not be easily discernible in the original input space, mitigating issues as eigenvalue multiplicities. To be specific, let $\boldsymbol{\lambda}$ be the column vector of graph normalized Laplacian eigenvalues, we process each of them using the Mexican hat kernel (as illustrated in Figure \ref{fig:spec_token}.B)
\begin{align}\label{eq:SpectralKernel}
	g(\theta_i\lambda_i) = \frac{2}{\sqrt{3} \pi^{1/4}} (1-\theta_i^2 \lambda_i^2) \exp(-\frac{\theta_i^2 \lambda_i^2}{2}),
\end{align} 
which a specific type of kernel used in signal analysis for feature extraction \cite{VertexFrequency2019}. It is optional to choose other functions referred to as \emph{spectral kernels} here like heat kernel, Gaussian kernel \cite{Djuric2018} or even trigonometric functions \cite{bo2023specformer}. It is also worth mentioning that the expansion of spectrum can be done either before the kernel or after the kernel. We then compute the following \emph{spectral attention} among the embedded eigenvalues through the Softmax with tunable weight matrix $W_1$:
\begin{align}\label{eq:SpectralAttention}
	\boldsymbol{s} = (s_j) = \Big( \frac{\exp (W_1 g(\boldsymbol{\theta} \cdot \boldsymbol{\lambda})_i  }{\sum_j \exp (W_1 g(\boldsymbol{\theta} \cdot \boldsymbol{\lambda})_j } \Big) 
\end{align}
and initialize a vector on the graph spectral token labeled by $0$ as 
\begin{align}\label{eq:SpectralAttention2}
	z_0^{(0)} = \boldsymbol{s} \odot W_2 \boldsymbol{\lambda},
\end{align}
where $\odot$ means Hardmard product on vectors.

\subsection{SubFormer-Spec Architecture}

In SubFormer \cite{pengmei2023transformers}, the shallow MP-GNN updates ordinary graph features followed by a standard Transformer which utilizes the self-attention mechanism to learn the coarse-grained tree \cite{JunctionTree2018,HIMP2020}. Consider two matrices: $U_G \in \mathbb{R}^{n \times n}$ and $U_T \in \mathbb{R}^{m \times n}$. The columns of $U_G$ and $U_T$ represent the Laplacian eigenvectors of a graph and its coarse-grained tree, respectively. Additionally, let $S \in {0,1}^{m \times n}$ denote the matrix that assigns graph nodes to the corresponding nodes of its coarse-grained tree.

We first concatenate $U_T$ and the matrix product $S U_G$, processed through learnable functions $\Phi_i$, as the eigenvector positional encoding (EPE):
\begin{align}
	U = [\Phi_1 (U_T), \Phi_2 (S U_G)]
\end{align}
Learnable functions can be implemented as either a fully-connected layer or as a SignNet, which is designed to ensure eigenvector invariance to sign flips, as discussed in \cite{lim2022sign}. Let $Z \in \mathbb{R}^{m \times d}$ denote the coarse-grained tree feature. We prepare the input for the Transformer block by concatenating $Z$ with node degree positional embedding $D_T$ (DPE) and the previously mentioned EPE.
\begin{align}
	Z^{'(0)} = \text{FFL}([Z, D_T]), \quad  
	Z^{(0)} = \text{FFL}([Z^{'(0)}, U]).
\end{align}
Note that we set $Z^{(0)}_0 = z_0^{(0)}$ defined in Eq.\eqref{eq:SpectralAttention2} which incorporates the spectral information of \emph{both} $G$ and $T$. Then EPE is optional in our architecture since both the graph spectral token and MP-GNN layers already provide sufficient structural information. Then We update $Z^{(0)}$ using the standard Transformer.

\begin{algorithm}
	\caption{SubFormer-Spec}
	\begin{algorithmic}[2]
		\State $X = \text{Embedding}(X_{ir}) \in \mathbb{R}^{n \times d_1}, Z= \text{Embedding}(Z_{js}) \in \mathbb{R}^{m \times d_2}$
		\Comment{Input embedding}
		\State $X'^{(l} =  \text{MPNN}(X^{l} )$.
		\State $X^{(l+1)} = X^{(l)'} + \text{FFL}( S^T Z^{(l)} W_2^{(l)})$. \Comment{Expanding coarse-grained feature to the original graph}
		\State $Z^{(l+1)} = Z^{(l)} + \text{FFL} (S X^{(l+1)} W_3^{(l)})$. \Comment{Compressing graph feature to the coarse-grained tree}
		
		\State $D_T = (d_{T,jr}) \in \mathbb{R}^{m \times d_T}$ \Comment{Embedding of node degrees}
		\State $Z^{(0)} = \text{FFL}([Z^{(0)}, D_T]) \in \mathbb{R}^{n \times d}$ \Comment{Concatenation with DPE}
		\If{initialize message-passing with EPE}
		\State $U = [\Phi_1(U_T), \Phi_2(U_G)]$
		\Comment{Preparing EPE using a Fully-connected layer or SignNet}
		\State $Z^{(0)} = \text{FFL}([Z^{(0)}, U]) \in \mathbb{R}^{n \times d}$  \Comment{Concatenation with EPE}
		\EndIf
		
		\State $\boldsymbol{\lambda} = [\boldsymbol{\lambda}_T, \boldsymbol{\lambda}_G], \boldsymbol{\theta} = (\theta_i), W_1, W_2$ \Comment{Initialization of spectral input with three collections of weights}
		\State $g(\theta_i\lambda_i) = (2/\sqrt{3} \pi^{1/4}) (1-\theta_i^2 \lambda_i^2) \exp(-\theta_i^2 \lambda_i^2/2)$ \Comment{Feature Extraction Using Mexican Kernel Function} 
		\State $\boldsymbol{s} = \text{Softmax}(W_1  g(\boldsymbol{\theta} \cdot \boldsymbol{\lambda})), z_0^{(0)} = \boldsymbol{s} \odot W_2 \boldsymbol{\lambda}$
		\Comment{High dimensional embedding weighted by correlation scores}
		\State Transformer Encoder
		\State Read out the auxiliary token with three-layer MLPs
	\end{algorithmic}
\end{algorithm}

\section{Emperical Results}
\label{sec:others}

\subsection{ZINC and Long-Range Graph Benchmark Datasets}
The ZINC dataset \cite{irwin2012zinc} serves as a standard benchmark consisting of small drug-like molecular graphs. Incorporating spectral information into the SubFormer-Spec model has shown to slightly improve performance, making it comparable to other state-of-the-art methods. It's noteworthy that many contemporary graph learning methods perform equally well on the ZINC dataset. For instance, SubFormer-Spec demonstrates a lower validation error than some methods, while achieving a similar test error, as detailed in Table \ref{tab:zinc}. 

In the domain of long-range graph benchmarks, particularly with the Peptides-Struct and Peptides-func datasets \cite{dwivedi2022long}, SubFormer-Spec outperforms its predecessor, the original SubFormer model as listed in Table \ref{tab:lrgb}. Graphs in these datasets are substantially larger than those in the ZINC dataset, incorporating a greater degree of long-range interactions. Interestingly, the benefit of incorporating the spectral token is more prominent in large graph datasets. This aligns with the findings in \cite{wilson2008study}, which indicate that the eigen spectrum of graphs becomes an increasingly powerful discriminative feature as graph size escalates.

\begin{table}[ht]
	\centering
	\caption{Results on ZINC dataset. Top 3 results are highlighted.Available benchmarks are taken from \cite{rampavsek2022recipe,he2023generalization,yang2023towards,bo2023specformer,yang2022spectrum}}
    \label{tab:zinc}
	\vspace{3mm}
	\centering
	\begin{tabular}{@{}cccc@{}}
		\toprule
		\textbf{Model} & \textbf{Test MAE($\downarrow$)} & \textbf{Valid MAE} & \textbf{Walltime(s)} \\
		\midrule
		GCN            & 0.278±0.003            & -                & 4        \\
        GAT            & 0.384±0.007          & - &    13 \\
		GIN            & 0.526±0.051             & -            & 10            \\
		HIMP           & 0.151±0.006            & -            & -            \\
  \midrule
		Transformer+LapPE             & 0.226±0.014            & -          & 45              \\
		SAN            & 0.181±0.004            & -             & 74           \\
        SAN+RWPE            & 0.104±0.004            & -             & 135           \\
		Graphormer     & 0.122±0.006            & -              & -          \\
		GraphGPS       & 0.070±0.004            & -  & 21                      \\
    \midrule
		Graph MLP-mixer& 0.073±0.001            & -                 & 6       \\
		Graph Vit      & 0.085±0.005            & -             & -           \\
    \midrule
		Specformer     & \textbf{0.066±0.003}   & - & 156                       \\
		PDF            & \textbf{0.066±0.002}   & 0.085±0.004    & 4          \\
        Spec-GN      & 0.070±0.002 & 0.088±0.003 & - \\
  \midrule
		SubFormer      & 0.077±0.003            & 0.085±0.002     & 3         \\
		SubFormer-Spec & \textbf{0.068±0.005}   & \textbf{0.072±0.003} &  7   \\
		\bottomrule
	\end{tabular}
\end{table}

\begin{table}[ht]
	\centering
	\caption{Results on Peptides-Struct/Func datasets from long-range graph benchmarks. Top 3 results are highlighted. Available benchmarks are taken from \cite{rampavsek2022recipe,he2023generalization,huang2023growing}. }
    \label{tab:lrgb}
	\vspace{3mm}
	\begin{minipage}{.5\linewidth}
		\centering
		\begin{tabular}{@{}cc@{}}
			\toprule
			\textbf{Peptides-Func} & \textbf{Test AP($\uparrow$)} \\
			\midrule
			GCN            & 0.5930±0.0023                \\
			GINE            & 0.5498±0.0079                \\
   HIMP & 0.5672±0.0038 \\
   GTR & 0.6519±0.0036 \\
   \midrule
			Transformer+LapPE             & 0.6326±0.0126                \\
   			SAN+LapPE            & 0.6384±0.0121                \\
			SAN+RWPE            & 0.6439±0.0075                \\
			GraphGPS       & 0.6535±0.0041                \\
   GraphTrans & 0.6313±0.0039 \\
   \midrule
			Graph MLP-mixer& \textbf{0.6970±0.0080}                \\
			Graph Vit      & 0.6942±0.0075       \\
   \midrule
			SubFormer      &   0.6732±0.0045     \\
                GraphTrans-Spec & \textbf{0.6957±0.0115} \\
			SubFormer-Spec   & \textbf{0.7014±0.0086}\\
			\bottomrule
		\end{tabular}
	\end{minipage}%
	\begin{minipage}{.5\linewidth}
		\centering
		\begin{tabular}{@{}ccc@{}}
			\toprule
			\textbf{Peptides-Struct} & \textbf{Test MAE($\downarrow$)} & \textbf{Walltime(s)} \\
			\midrule
			GCN            & 0.3496±0.0013      &5          \\
			GINE            & 0.3547±0.0045     &4           \\
            HIMP & 0.2653±0.0010 & 6 \\
            GTR & 0.2502±0.0017 & -\\
   \midrule
			Transformer+LapPE             & 0.2529±0.0016      &6          \\
   			SAN+LapPE            & 0.2683±0.0043      &54          \\
			SAN+RWPE            & 0.2545±0.0012      &50          \\
			GraphGPS       & 0.2500±0.0005     &12           \\
            GraphTrans & 0.2777±0.0025 & - \\
    \midrule
			Graph MLP-mixer& 0.2475±0.0015    &9            \\
			Graph Vit      & \textbf{0.2449±0.0016}  &9     \\
   \midrule
			SubFormer      & \textbf{0.2464±0.0012}  &7     \\
                GraphTrans-Spec & 0.2487±0.0009 & 7\\
			SubFormer-Spec   & \textbf{0.2441±0.0011} & 7\\
			\bottomrule
		\end{tabular}
	\end{minipage}
\end{table}

\subsection{MoleculeNet} 
For a comprehensive evaluation, we selected several datasets from MoleculeNet, adhering to the recommended settings outlined in \cite{wu2018moleculenet}. Table \ref{tab:moleculenet} shows the effectiveness of the spectral token across a diverse range of chemical datasets. Despite kernel methods are generally doing better on small-scale datasets as SIDER and BBBP. 

\begin{table*}[h]
	\caption{Results on MoleculeNet. Top 3 results are highlighted.Available benchmarks are taken from \cite{wu2018moleculenet,HIMP2020,Graphormer2021, pengmei2023transformers}}
	\vspace{3mm}
	\label{tab:moleculenet}
	\centering
	\small 
	\begin{tabular}{@{}ccccccc@{}}
		\toprule
		\textbf{Dataset}             & \textbf{TOX21}                      & \textbf{TOXCAST}              & \textbf{MUV}                    & \textbf{HIV} & \textbf{SIDER} & \textbf{BBBP}                  \\
		\midrule
		Num. Task           & 12                         & 617                  & 17                     & 1  & 27 & 1                  \\
		Metric               & ROC-AUC($\uparrow$) & ROC-AUC              &  RPC-AUC($\uparrow$)                           & ROC-AUC  & ROC-AUC & ROC-AUC     \\ 
		\midrule
		RF                  & 0.769±0.015                & -                    & -                      & 0.684±0.009 & \textbf{0.714±0.000}  & \textbf{0.781±0.006}    \\
		XGBoost             & 0.794±0.014                & 0.640±0.005          & 0.086±0.033            & 0.756±0.000 & 0.656±0.027           & 0.696±0.000             \\
		Kernel SVM          & 0.822±0.006                & 0.669±0.014          & \textbf{0.137±0.033}   & 0.792±0.000 & \textbf{0.682±0.013} & \textbf{0.729±0.000}\\
		LR                  & 0.794±0.015                & 0.605±0.003          & 0.070±0.009            & 0.702±0.018 & 0.643±0.011           & 0.699±0.002             \\
  \midrule
		GCN                 & 0.840±0.004                & 0.735±0.002          & 0.114±0.029            & 0.761±0.010 & 0.601±0.013         & 0.712±0.012                       \\
		GIN                 & 0.850±0.009                & \textbf{0.741±0.004}          & 0.091±0.033            & 0.756±0.014  & 0.571±0.012   & 0.689±0.013                       \\
		HIMP                & \textbf{0.874±0.005}       & 0.721±0.004          & 0.114±0.041            & 0.788±0.080  & 0.562±0.013                    & 0.701±0.011                       \\

  \midrule
        Graphormer & - &- &- & \textbf{0.805±0.005} &- &-\\
		GraphGPS            & 0.841±0.003                & 0.714±0.006          & 0.087                  & 0.788±0.010  & 0.607±0.011           & 0.651±0.038             \\
        \midrule
		SchNet              & 0.769                      & 0.685                & -                      & -            & 0.597                 & -                       \\
		EGNN                & \textbf{0.854}             & 0.739                & -                      & -            & 0.604                 & -                       \\
		ClofNet             & 0.842                      & 0.700                & -                      & -            & 0.603                 & -                       \\
        \midrule
		SubFormer           & 0.851±0.008                & \textbf{0.752±0.003} & \textbf{0.182±0.019}   & \textbf{0.795±0.008} & 0.678±0.014             &0.703±0.010                        \\
		SubFormer-Spec      & \textbf{0.867±0.005}       & \textbf{0.764±0.005} & \textbf{0.203±0.012}   & \textbf{0.805±0.008} & \textbf{0.683±0.005} & \textbf{0.731±0.021} \\ 
		\bottomrule
	\end{tabular}
\end{table*}

\subsection{Organic Photovoltaics with Donor-Acceptor (OPDA) Structure Dataset.}

Graph transformers, akin to their text counterparts, excel in addressing long-range interactions in large graphs, a task particularly relevant in chemical contexts where charge-transfer is a critical long-range phenomenon. In OPDA molecules \cite{landis2012computational}, charge-transfer involves a delocalizing process where an electron is transferred from a localized donor group to a distant acceptor group. These molecules, designed with separated donor and acceptor groups, are pivotal in tuning physical properties such as the energy gap between the highest occupied molecular orbital (HOMO) and lowest unoccupied molecular orbital (LUMO).

\begin{table}
	\centering
	\caption{Results on OPDA dataset. Top 1 results are highlighted. All results are MAE, lower the better. Each model is allowed to train 300 epochs.}
         \small
        \label{tab:OPDA}
	\vspace{5pt}
	\begin{tabular}{@{}ccccccc@{}}
		\toprule
		\textbf{Property} & \textbf{$\text{E}_{\text{HOMO-LUMO}}$} & \textbf{Packing density} & \textbf{$\text{E}_{\text{HOMO}}$} & \textbf{$\text{E}_{\text{LUMO}}$} & \textbf{Dip. moment} & \textbf{Walltime(s/epoch)} \\
		\midrule
		GCN               & 0.845                       & 0.462                    & 1.526                & 0.647                & 1.369          &0.94                \\
		GIN               & 0.210                       & 0.073                    & 0.273                & 0.557                & 1.236          &0.91                \\
		GATv2             & 0.299                       & 0.069                    & 1.357                & 0.445                & 1.548    &1.04                      \\
		GraphGPS &0.080&0.015&0.068&0.044&0.966 & 1.92 \\
		\midrule
		SchNet            & 0.331                       & 0.215                    & 0.429                & 0.297                & 1.236   & 1.58                       \\
		DimeNet++         & 0.139                       & 0.026                    & 0.190                & 0.093                & 1.089      &3.97                    \\
		ClofNet           & 0.132                       & 0.016                    & 0.123                & 0.083                & 1.259          & 1.34                \\
		EGNN              & 0.112                       & 0.174                    & 0.138                & 0.076                & 1.216         & 1.14                 \\
		\midrule
		SF         & 0.070              & 0.009           & 0.050       & 0.031       & 0.997  & 1.53               \\
		SF-Spec    & \textbf{0.047}              &\textbf{0.008}           & \textbf{0.043}       & \textbf{0.028}       & \textbf{0.888} & 1.61 \\ 
		\bottomrule              
	\end{tabular}
\end{table}

\newpage
\bibliographystyle{unsrt}  
\bibliography{templateArxiv.bib}

\appendix

\newpage
\section{Datasets}
\begin{figure}[ht]
    \centering
    \includegraphics[width=\textwidth]{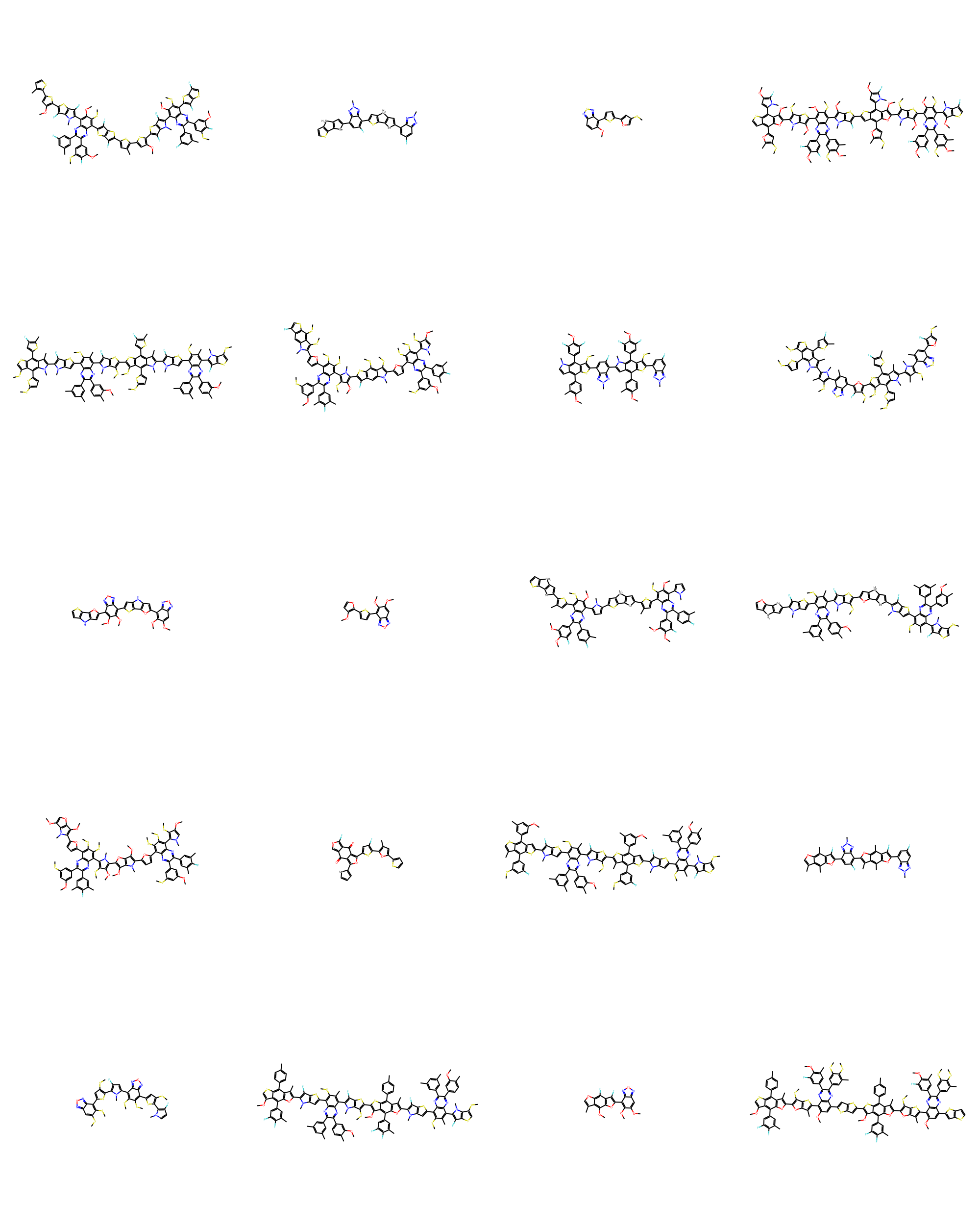}
    \caption{Illustration of molecules included in the OPDA dataset, samples are randomly drawn.}
    \label{fig:opda_mols}
\end{figure}

We summarize the datasets used in the study, which are mainly composed of MoleculeNet datasets \cite{wu2018moleculenet} (TOX21, TOXCAST, MUV, HIV, SIDER and BBBP), ZINC \cite{irwin2012zinc}, long-range graph benchmarks (Peptides-func/struct) \cite{dwivedi2022long}, and OPDA dataset \cite{landis2012computational}. We follow all recommended data split methods and evaluation metrics as proposed in the original literature. For OPDA dataset, we randomly split the dataset 8:1:1 for training, validation, and testing. Additionally, we illustrates random samples from OPDA datasets to showcase the characteristics of OPDA systems in Figure \ref{fig:opda_mols}.
\begin{table}[ht]
\centering
\small 
\caption{Summary of dataset statistics used in this study.} 
\label{tab:stats}
\begin{tabular}{p{2.5cm}p{1.2cm}p{1.5cm}p{1.5cm}p{1.5cm}p{2.5cm}p{2.5cm}}
\toprule
\textbf{Dataset} & \textbf{\# Graphs} & \textbf{Avg. Nodes} & \textbf{Avg. Cliques} & \textbf{Avg. Edges} & \textbf{Task} & \textbf{Metric} \\
\midrule
ZINC & 12000 & 23.2 & 14.2 & 24.9 & Regression & Mean Abs. Error \\
Peptides-func & 15535 & 150.9 & 168.9 & 307.3 & 10-task classif. & Avg. Precision \\
Peptides-struct & 15535 & 150.9 & 168.9 & 307.3 & 11-task regression & Mean Abs. Error \\
OPDA & 5356 & 55.8 & 28.1 & 63.9 & Regression & Mean Abs. Error \\
TOX21 & 7831 & 18.6 & 12.9 & 38.6 & 12-task classif. & ROC-AUC \\
TOXCAST & 8597 & 18.7 & 13.0 & 38.4 & 617-task classif. & ROC-AUC \\
MUV & 93087 & 24.2 & 14.1 & 52.6 & 17-task classif. & RPC-AUC \\
HIV & 41127 & 25.5 & 15.8 & 54.9 & 1-task classif. & ROC-AUC \\
SIDER & 1427 & 33.6 & 25.9 & 70.7 & 27-task classif. & ROC-AUC \\
BBBP & 2050 & 23.9 & 14.5 & 51.6 & 1-task classif. & ROC-AUC \\
\bottomrule
\end{tabular}
\end{table}

\newpage
\section{SubFormer-Spec Hyperparameters}

The hyperparameters applied to all benchmark datasets in this study are detailed in Table \ref{tab:hps_combined_1} and \ref{tab:hps_combined_2}. We arbitrarily choose to use 16 eigenvalues from the original graph and coarse-grain tree (16+16=32 total) for samll graphs, which should be treated a vital hyper-parameter theoretically. And we picked 32+32 eigenvalues for Peptides-Struct  dataset and 64+32 for Peptides-Func dataset. We did not perform a systematic search of hyperparameters due to limited computational resources and we are already satisfied with the performance. We use common random seeds as 4321, 1234, 42, 1, etc.

\begin{table}[ht]
\centering
\caption{Hyperparameter settings for ZINC and  Long-range Graph Benchmarks }
\vspace{3mm}
\label{tab:hps_combined_1}
\begin{tabular}{cccc}
\toprule
\textbf{Model Component} & \textbf{ZINC} & \textbf{Peptides-Struct}& \textbf{Peptides-Func} \\
\midrule
Optimization & & & \\
\quad Warmup Epoch & 50 & 0 & 20 \\
\quad Epoch & 950 & 100 & 200\\
\quad Learning rate & 0.001 & 0.0005 & 0.001 \\
\quad Optimizer & AdamW & AdamW & AdamW \\
\quad Scheduler & Cosine & ROP& Cosine \\
\quad Batch size & 32 & 64& 64 \\
\midrule
Local MP & & & \\
\quad \# Layers & 2 & 2 & 2\\
\quad \# Hidden Features & 64 & 64& 64 \\
\quad MP Type & GINE & GINE & GINE\\
\quad Aggregation & Sum & Sum & Sum\\
\quad Activation & ReLU & ReLU & ReLU\\
\quad Tree Activation & LeakyReLU & LeakyReLU & LeakyReLU\\
\quad Dropout & 0 & 0.05 & 0.05\\
\quad Edge Dropout & 0 & 0 & 0\\
\midrule
PE & & \\
\quad PE Dim. & 10 & - & -\\
\quad PE Type & DEG, LapPE & DEG, None & DEG, None\\
\quad PE Merge & Concat & Concat  & Concat\\
\midrule
Transformer & & \\
\quad \# Hidden Features & 128 & 128 & 128\\
\quad \# FFN Hidden Features & 128 & 128 & 128\\
\quad \# Layers & 3 & 3 & 3 \\
\quad \# Heads & 8 & 8 & 8 \\
\quad Activation & ReLU & ReLU & GELU\\
\quad Dropout & 0.1 & 0.05 & 0.2\\
\midrule
Readout & & \\
\quad \# Hidden Features & 192 & 128 & 128\\
\quad Activation & ReLU & ReLU & GELU \\
\bottomrule
\end{tabular}
\end{table}

\begin{table}[ht]
\centering
\small
\caption{Hyperparameter settings for TOX21, TOXCAST, MUV, HIV, SIDER, and BBBP}
\label{tab:hps_combined_2}
\begin{tabular}{ccccccc}
\toprule
\textbf{Model Component} & \textbf{TOX21} & \textbf{TOXCAST} & \textbf{MUV} & \textbf{HIV} & \textbf{SIDER} & \textbf{BBBP} \\
\midrule
Optimization & & & & & & \\
\quad Epoch & 50 & 50 & 50 & 15 & 50 & 100 \\
\quad Learning rate & 0.0001 & 0.0001 & 0.0001 & 0.0001 & 0.0005 & 0.0005 \\
\quad Optimizer & AdamW & AdamW & AdamW & AdamW & AdamW & AdamW \\
\quad Scheduler & None & None & None & ROP & ROP & ROP \\
\quad Batch size & 32 & 32 & 128 & 32 & 32 & 64 \\
\midrule
Local MP & & & & & & \\
\quad \# Layers & 6 & 3 & 3 & 4 & 3 & 2 \\
\quad \# Hidden Features & 256 & 256 & 256 & 128 & 256 & 64 \\
\quad MP Type & GINE & GINE & GINE & AGAT & GINE & GINE \\
\quad Aggregation & Sum & Sum & Sum & Sum & Sum & Sum \\
\quad Activation & ReLU & ReLU & ReLU & ReLU & ReLU & ReLU \\
\quad Tree Activation & LeakyReLU & LeakyReLU & LeakyReLU & LeakyReLU & LeakyReLU & LeakyReLU \\
\quad Dropout & 0.1 & 0.2 & 0.1 & 0 & 0.2 & 0 \\
\quad Edge Dropout & 0.1 & 0.2 & 0 & 0 & 0.2 & 0 \\
\midrule
PE & & & & & & \\
\quad PE Dim. & 16 & 20 & 16 & 16 & 10 & 10 \\
\quad PE Type & DEG, - & DEG, - & DEG, - & DEG, LapPE & DEG, LapPE & DEG, LapPE \\
\quad PE Merge & Concat & Concat & Concat & Concat & Concat & Concat \\
\midrule
Transformer & & & & & & \\
\quad \# Hidden Features & 512 & 512 & 512 & 256 & 256 & 128 \\
\quad \# FFN Hidden Features & 1024 & 512 & 512 & 512 & 256 & 128 \\
\quad \# Layers & 4 & 4 & 4 & 4 & 4 & 3 \\
\quad \# Heads & 8 & 8 & 8 & 8 & 8 & 4 \\
\quad Activation & ReLU & ReLU & ReLU & ReLU & ReLU & ReLU \\
\quad Dropout & 0.2 & 0.5 & 0.2 & 0.3 & 0.5 & 0.2 \\
\midrule
Readout & & & & & & \\
\quad \# Hidden Features & 768 & 768 & 768 & 256 & 768 & 128 \\
\quad Activation & ReLU & ReLU & ReLU & ReLU & ReLU & ReLU \\
\bottomrule
\end{tabular}
\end{table}

\textbf{OPDA Dataset.} We allow all models to train 300 epoches. All targets are normalized and we only consider heavy atoms following the convention. For OPDA tasks, we use a consistent architecture comprising three GINE MP-GNN layers with ReLU and LeakyReLU activations and four transformer encoder layers with 128 channels, 8 MHA heads, and 256-channel FFNs with GELU activation. A dropout rate of 0.05 is applied for regularization. Positional encoding is exclusively DEG. The readout MLP is 128-channel wide with GELU activation and we just readout from the coarse-grained graph. The model is trained with batches of 64 using an Adam optimizer (learning rate 0.001) and an ROP scheduler. 

For all MP-GNNs, we set the dimension to 128-channel wide and 8 layers with a dropout rate of 0.2. For SchNet, we use the default hyperparameters as implemented in the pytorch geometric package. For DimeNet++, we use hidden channel width of 128, 3 interaction blocks, embedding size of 64. For EGNN and ClofNet, we set the hidden dimension to 128 with 4 layers with the attention. While the optimizer settings are the same.

For GraphGPS, we apply 6 layers with multihead attention, with the random walk position encoding dimension of 20, with the dropout rate of 0.2, and the summation aggregation. We use AdamW optimizer with learning rate of 0.001 and the ROP scheduler. 

\clearpage

\section{Training Curves}
To better demonstrate the training dynamics of SubFromer-Spec model, we plotted the training curves in this section for visualization. 

\begin{figure}[ht]
    \centering
    \includegraphics[width=\textwidth]{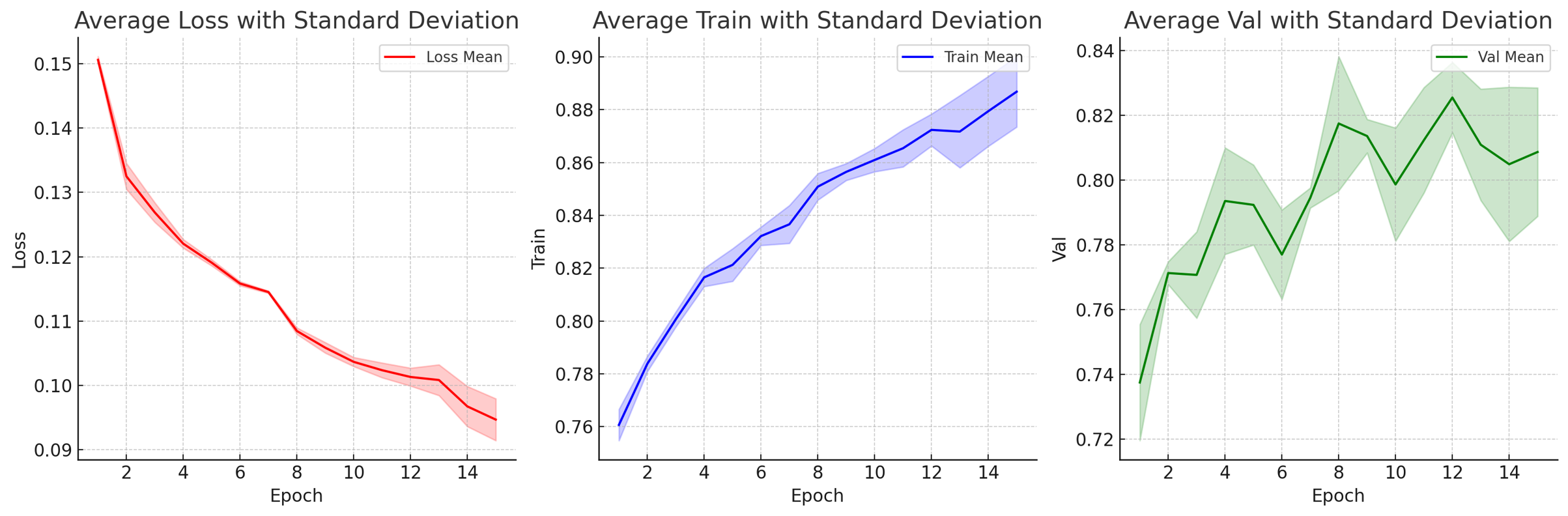}
    \caption{Training curve of the SubFormer-Spec on HIV dataset.}
    \label{fig:tr_hiv}
\end{figure}

\begin{figure}[ht]
    \centering
    \includegraphics[width=\textwidth]{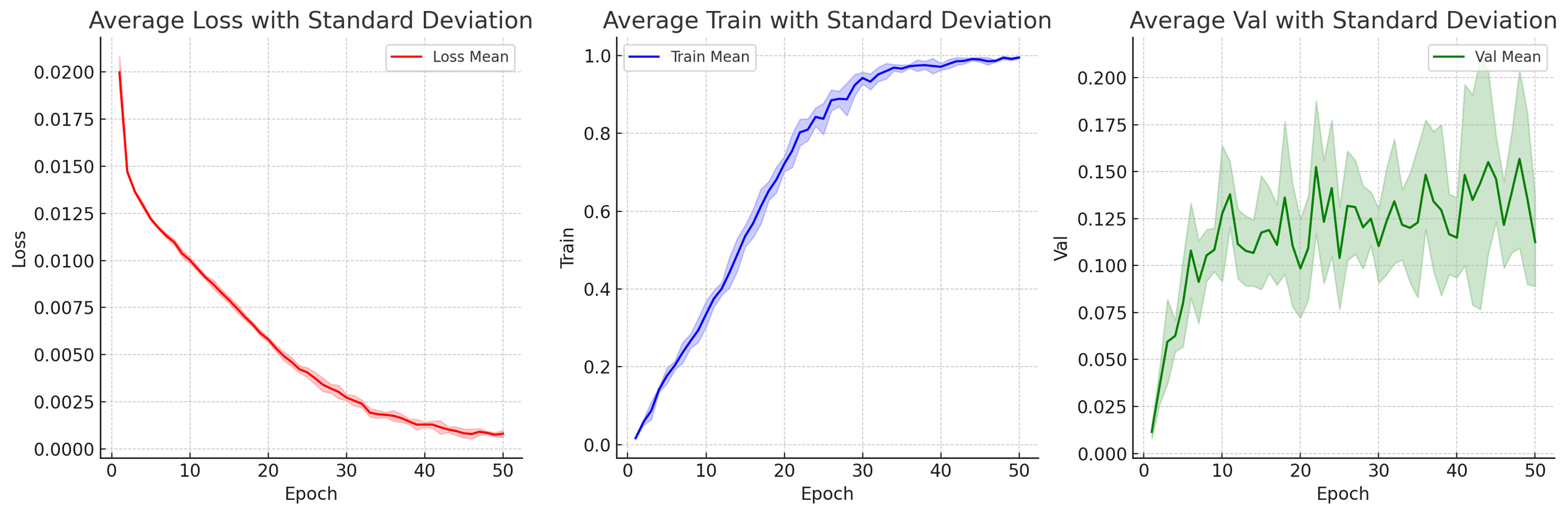}
    \caption{Training curve of the SubFormer-Spec on MUV dataset.}
    \label{fig:tr_muv}
\end{figure}

\begin{figure}[ht]
    \centering
    \includegraphics[width=\textwidth]{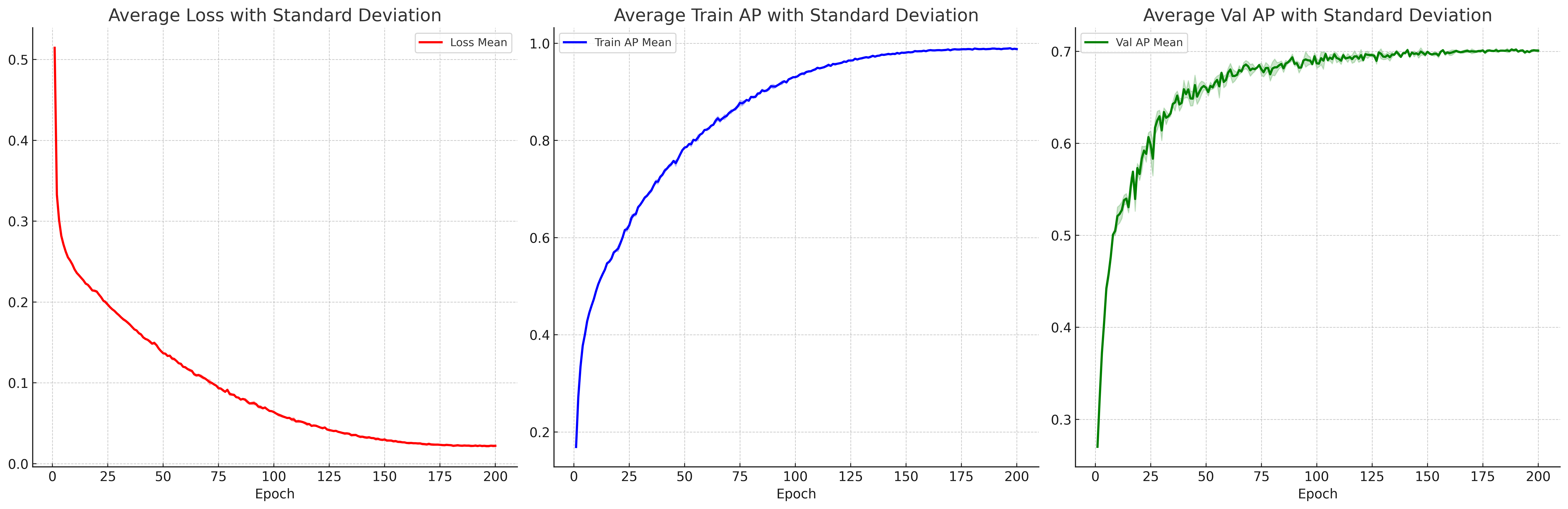}
    \caption{Training curve of the SubFormer-Spec on Peptides-Func dataset.}
    \label{fig:tr_pf}
\end{figure}

\begin{figure}[ht]
    \centering
    \includegraphics[width=0.67\textwidth]{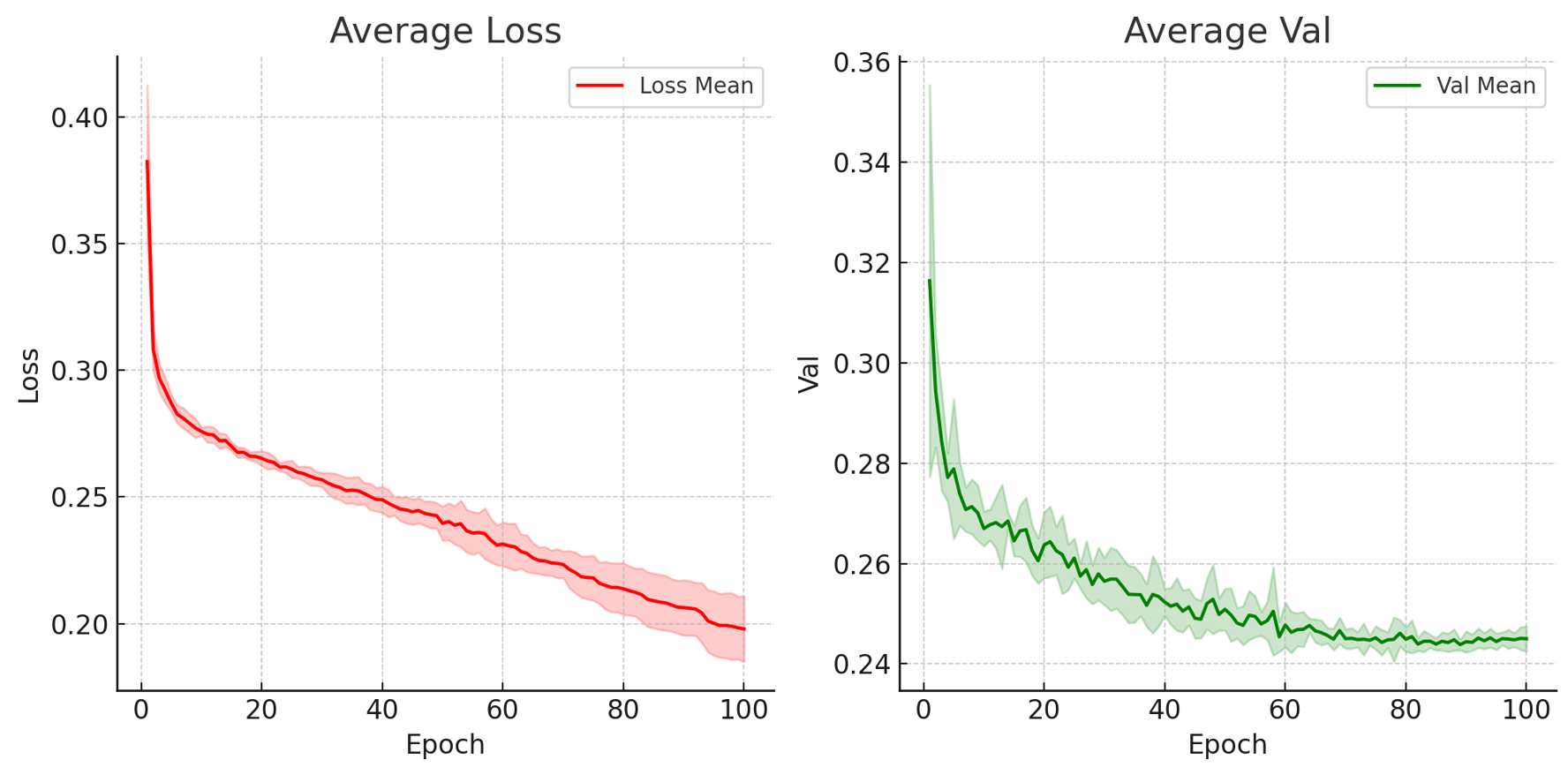}
    \caption{Training curve of the SubFormer-Spec on Peptides-Struct dataset.}
    \label{fig:tr_ps}
\end{figure}

\begin{figure}[ht]
    \centering
    \includegraphics[width=0.67\textwidth]{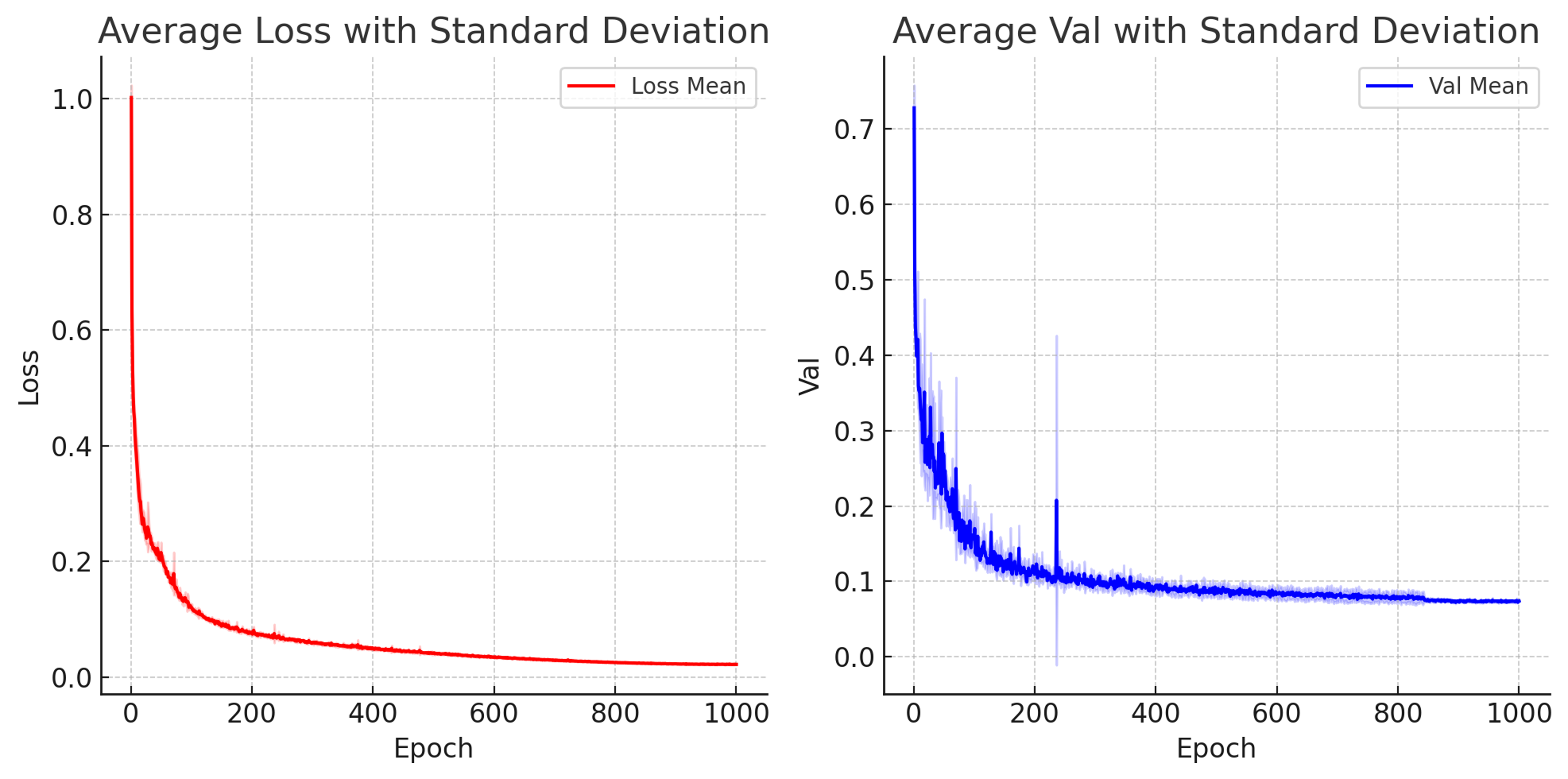}
    \caption{Training curve of the SubFormer-Spec on ZINC dataset.}
    \label{fig:tr_zinc}
\end{figure}

\begin{figure}[ht]
    \centering
    \includegraphics[width=\textwidth]{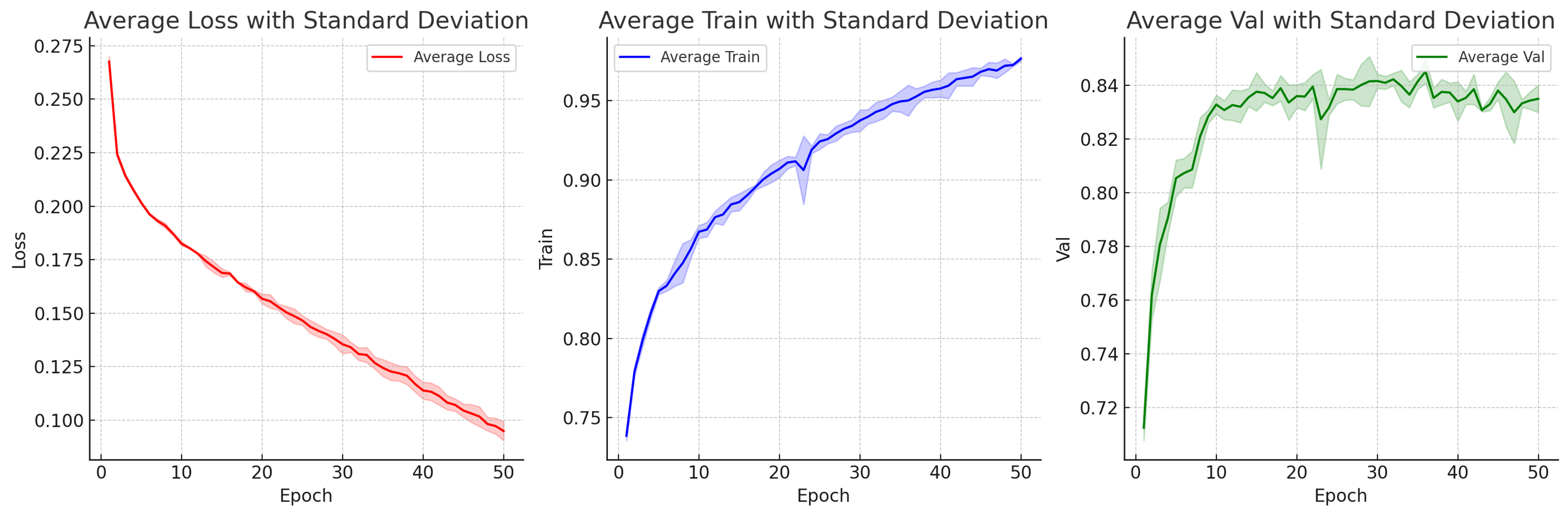}
    \caption{Training curve of the SubFormer-Spec on TOX21 dataset.}
    \label{fig:tr_tox21}
\end{figure}

\begin{figure}[ht]
    \centering
    \includegraphics[width=\textwidth]{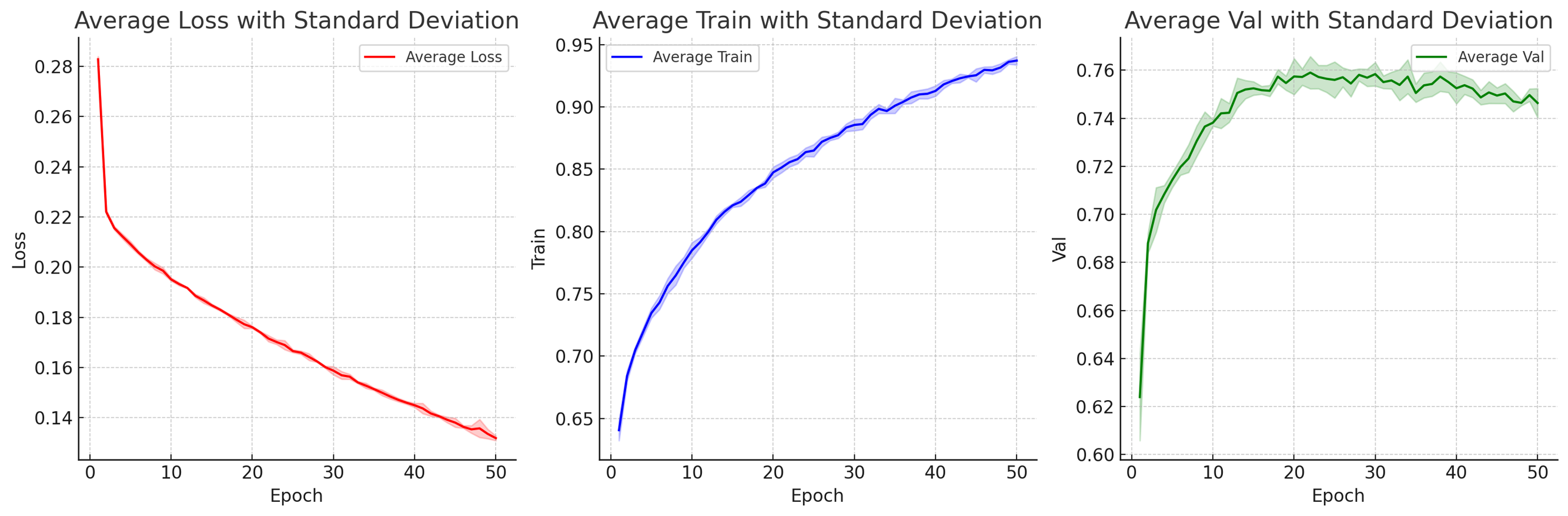}
    \caption{Training curve of the SubFormer-Spec on TOXCAST dataset.}
    \label{fig:tr_toxcast}
\end{figure}

\newpage

\clearpage

\end{document}